\documentclass{article}
\usepackage{iclr2025_conference,times}
\usepackage{amsmath,amsfonts,bm}

\def\eqref#1{equation~\ref{#1}}
\def\1{\bm{1}}

\DeclareMathAlphabet{\mathsfit}{\encodingdefault}{\sfdefault}{m}{sl}
\SetMathAlphabet{\mathsfit}{bold}{\encodingdefault}{\sfdefault}{bx}{n}

\usepackage{amsmath,amssymb}
\usepackage{graphicx}
\usepackage{booktabs}
\usepackage{subcaption}
\usepackage{hyperref}
\usepackage{url}
\iclrfinalcopy

\renewcommand{\eqref}[1]{(\ref{#1})}

\title{MedICL: Modality-Anatomy-Aware Visual In-Context Learning for Medical Visual Question Answering}

\author{Ji-jun Park, Soo-joon Choi, Seungmin Han, Haeun Kwon, Jiwon Jeong, Taeyang Yoon \\
        Dongguk University, South Korea \\
        \texttt{taeyang.yoon@dongguk.edu}
}

\begin{document}

\maketitle

\begin{abstract}
Training-free adaptation of medical vision-language models (LVLMs) is a deployment requirement, yet visual in-context learning (ICL) injects modality- and style-mismatched demonstrations into medical contexts, teaching a frozen model the wrong answering style. We present MedICL, which curates demonstrations instead of retraining the model. MADS selects modality--anatomy aligned demonstrations that cover complementary interpretation dimensions; W2C curates them with weak retrieval signals and a strong verification vote that checks demonstrations rather than the model's own outputs; ECC answers with $W$ anchored demonstration sets, aggregates by abnormality-aware signatures, and re-answers once on conflict. On VQA-RAD, SLAKE, and PathVQA under the LLaVA-Med zero-shot protocol, MedICL improves the strongest zero-shot baseline by $+11$--$+14$ points on open-ended and $+7$--$+14$ on closed questions, recovers 71\% of fine-tuned open-ended performance on VQA-RAD at zero training cost, and stays robust to shifted or corrupted demonstration banks.
\end{abstract}

\section{Introduction}

Large vision-language models (LVLMs) have rapidly expanded what machines can do with images and text, from open-ended perception to interactive world modeling~\citep{xiong2026actworld,liu2023visual}. In medicine, this progress has produced assistants that answer questions about radiology and pathology images~\citep{li2023llavamed,moor2023medflamingo}. Medical visual question answering (Med-VQA) is arguably the most deployable form of this progress: a clinician points at an image, asks a question, and expects a short, usable answer. Yet deploying a single medical LVLM across departments remains impractical. Radiology, pathology, and multi-modal wards differ in modality, anatomy, and answer style; fine-tuning per department requires costly annotation campaigns that must be repeated whenever guidelines or equipment change. Training-free adaptation is therefore not a convenience but a requirement for scalable clinical deployment.

Visual in-context learning (ICL) is the most natural training-free channel: instead of updating parameters, we place a handful of \emph{image--question--answer} demonstration pairs into the context and let the frozen model imitate them. \citet{zhou2024visual} showed that how demonstrations are organized strongly affects the visual generalization of LVLMs, and recent work on multimodal ICL likewise reports that performance is highly sensitive to the quality and coverage of selected demonstrations~\citep{cheng2026mag}. However, existing demonstration selection was designed for natural images and generic captions. Medical imaging breaks its core assumption that ``similar means helpful''. A chest CT is globally similar to a chest X-ray, but the answer styles of the two modalities differ; a normal screening image is similar to an abnormal one, yet demonstrates the wrong interpretation granularity. Naive Top-$K$ retrieval therefore injects demonstrations that teach the frozen model the wrong answering style---a failure mode specific to medical ICL that motivates curation rather than retrieval alone.

A second obstacle is that no single component can reliably judge demonstration quality. Retrieval scores measure relevance, not answerability; letting the strong LVLM pick its own demonstrations invites self-preference, as the model tends to favor demonstrations that confirm its existing biases. Work on weak-to-strong generalization has shown that weak supervision can steer strong models across capabilities, but this division of labor has not been carried into inference-time demonstration curation. Meanwhile, medical LVLMs are known to need feedback anchored on abnormality perception~\citep{zhou2025improving}, yet that signal has only been used during fine-tuning. Sample curation at scale, studied for long-context alignment~\citep{si2025gateau}, similarly remains a training-time technology. In short: the three ingredients of reliable inference-time adaptation---structure-aware selection, weak-to-strong verification, and abnormality-grounded checking---exist in isolation but have never been composed into a single training-free pipeline for Med-VQA.

We propose \textbf{MedICL}, a training-free visual in-context learning framework that turns a frozen medical LVLM into a department-portable Med-VQA system by curating what it sees rather than retraining what it knows. MedICL contributes three modules. First, \textbf{MADS} (Modality-Anatomy-aware Demonstration Selection) compiles the query into typed slots (modality, anatomy, question type), filters the demonstration bank structurally, and selects a set that covers complementary interpretation dimensions with position-aware ordering. Second, \textbf{W2C} (Weak-to-Strong Demonstration Curation) splits the judging problem: cheap weak signals (a biomedical CLIP retriever and a lightweight answer classifier) shortlist candidates, and the strong LVLM verifies each candidate's answerability by checking whether it can reproduce the demonstration's own answer given the other demonstrations---a check on demonstrations rather than on the model's own outputs, which avoids self-preference. Third, \textbf{ECC} (Exemplar-Ensemble Consistency Checking) answers with $W$ demonstration sets anchored on different interpretation dimensions, extracts an abnormality-aware signature from each answer, aggregates by signature-weighted agreement, and triggers one bounded re-answer round when signatures conflict on abnormality presence.

Our contributions are:
\begin{itemize}
\item We identify demonstration curation as the central, previously unaddressed bottleneck of training-free medical ICL, and show that global-similarity retrieval systematically injects modality- and style-mismatched demonstrations into medical LVLM contexts.
\item We propose MedICL, the first training-free Med-VQA framework that composes structure-aware demonstration selection (MADS), weak-to-strong demonstration verification (W2C), and abnormality-grounded ensemble consistency (ECC) into a single inference-time pipeline, drawing the weak-to-strong principle into curation and the abnormality-aware principle of medical feedback~\citep{zhou2025improving} into aggregation.
\item We design a two-vote verification scheme whose verification target is the demonstration itself, which provably sidesteps the self-preference that plagues self-selecting ICL.
\item On VQA-RAD, SLAKE, and PathVQA under the LLaVA-Med zero-shot protocol, MedICL improves over the strongest zero-shot baseline by $+11$--$+14$ points on open-ended and $+7$--$+14$ points on closed questions, recovers $71\%$ of the fine-tuned open-ended performance on VQA-RAD at zero training cost, and remains robust to cross-corpus banks and corrupted labels.
\end{itemize}

\section{Related Work}

\subsection{Medical LVLMs and Med-VQA}
General-purpose LVLMs are adapted to biomedicine either by instruction tuning on curated image--text data~\citep{li2023llavamed,liu2023visual} or by few-shot multimodal prompting~\citep{moor2023medflamingo}. Discriminative Med-VQA models pre-train with masked vision--language objectives~\citep{li2023m2i2} or adapt CLIP to the medical domain~\citep{eslami-etal-2023-pubmedclip,zhang2023biomedclip}, achieving strong supervised accuracy but requiring per-dataset training. Benchmarks have evolved accordingly: VQA-RAD provides clinician-posed questions on radiology images~\citep{lau2018vqarad}, SLAKE adds modality, anatomy, and knowledge annotations~\citep{liu2022slake}, and PathVQA scales pathology QA to 32{,}799 pairs~\citep{he2020pathvqa}. Recent work continues this supervised line---noise-aware representation learning improves robustness under corrupted visual embeddings~\citep{pratama2026noise}, and grounded multimodal LLMs couple VQA with pixel-level segmentation for clinical trustworthiness~\citep{gu2026grounded}. All of these methods require gradient updates; MedICL is orthogonal and complementary, upgrading the zero-shot regime that they necessarily start from, and it can adopt their tuned encoders as stronger weak retrievers inside W2C.

\subsection{In-Context Learning and Demonstration Selection}
For text models, demonstration choice has been studied from retrieval~\citep{li-etal-2023-shot}, meta-learning~\citep{min-etal-2022-metaicl}, label bias~\citep{fei-etal-2023-mitigating}, active selection~\citep{margatina-etal-2023-active}, and shortcut analysis~\citep{tang-etal-2023-large} perspectives; retrieval-augmented VQA injects outside knowledge~\citep{lin-byrne-2022-retrieval}, context construction improves open-domain answering~\citep{su-etal-2023-context}, and long-context QA has been tackled with supervised contrastive learning~\citep{caciularu-etal-2022-long}. \citet{zhou2024visual} extended this line to LVLMs by organizing image--text demonstrations, and \citet{cheng2026mag} select demonstrations on a manifold in a learning-free manner for general multimodal tasks. \citet{si2025gateau} select influential samples for long-context alignment, but as a training-time technology. None of these criteria understands the modality--anatomy--question-type structure of medical imaging, and none verifies that a retrieved demonstration can actually be answered by the deployed model; MedICL addresses both gaps with structured submodular selection followed by two-vote verification.

\subsection{Weak-to-Strong Supervision and Contextual Faithfulness}
\citet{zhou2025weak} demonstrate that weak models can guide strong models toward multi-capability generalization, and synthetic-task reinforcement learning teaches LLMs contextual faithfulness~\citep{DBLP:conf/aaai/SiZGBWGLLHCQZCS26}, complementing context-faithful prompting at inference time~\citep{zhou-etal-2023-context}. Faithfulness of the evidence is equally central in high-stakes domain NLP, where timestamp-aware evidence determines whether predictions are trustworthy~\citep{hu2026ai}. In the medical domain specifically, \citet{zhou2025improving} show that abnormality-aware feedback is the signal that matters when improving medical LVLMs---but within a fine-tuning loop. MedICL transplants both ideas to inference time: weak retrievers and a weak answer classifier supply the first vote, the strong LVLM supplies a verification vote whose target is the demonstration (not its own generation), and abnormality perception arbitrates answer aggregation. This composition also connects to inference-time planning: just as long-horizon agents need an explicit global plan~\citep{si2026goal}, medical ICL needs an explicit plan over which interpretation dimensions the demonstrations must cover before any answer is produced.

\section{Method}
\label{sec:method}

\begin{figure}[!t]
  \centering
  \includegraphics[width=0.8\linewidth]{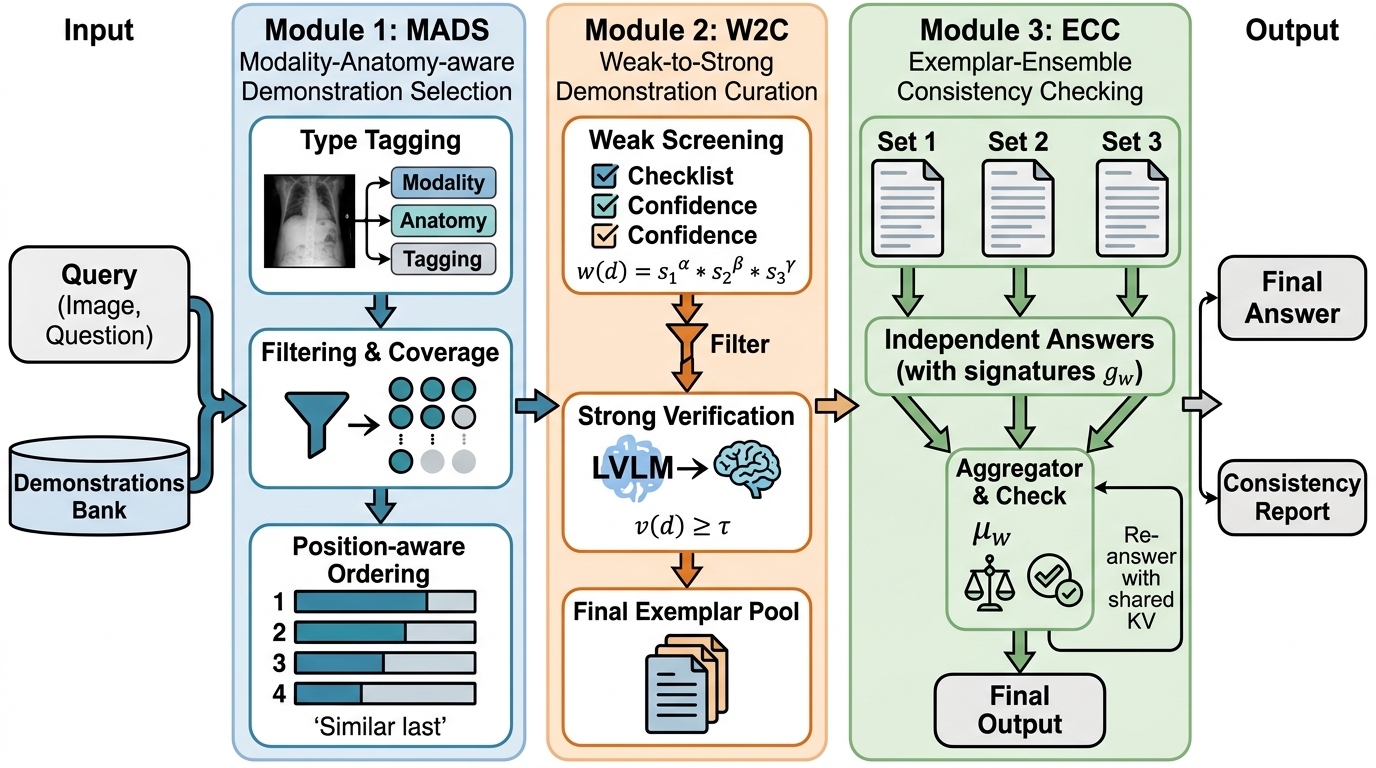}
  \caption{Overview of MedICL. A frozen medical LVLM answers each query through three training-free modules: MADS selects modality--anatomy aligned demonstrations with submodular coverage; W2C curates them with weak weak signals and a strong LVLM verification vote; ECC answers with $W$ anchored demonstration sets, aggregates by abnormality-aware signatures, and re-answers once on conflict.}
  \label{fig:arch}
\end{figure}

\subsection{Problem Formulation}

Let $\mathcal{B}=\{(I_j,q_j,a_j)\}_{j=1}^{N}$ be a bank of demonstration triples (image, question, answer) drawn from training corpora, and let $(I,q)$ be a query image--question pair. A frozen LVLM $f_\theta$ answers autoregressively conditioned on $W$ demonstration sets:
\begin{align}
p_\theta(a \mid I,q,\mathcal{D})=\prod_{t} p_\theta\!\left(a_t \mid a_{<t},\, q,\, I,\, \mathcal{D}\right),
\label{eq:icl}
\end{align}
where $\mathcal{D}\subset\mathcal{B}$ is the demonstration context, as formulated in Eq.~\eqref{eq:icl}. MedICL never updates $\theta$; it optimizes $\mathcal{D}$ instead, through three stages: MADS builds structured candidates (\S\ref{sec:mads}), W2C curates them (\S\ref{sec:w2c}), and ECC aggregates ensemble answers (\S\ref{sec:ecc}).

\subsection{MADS: Modality-Anatomy-aware Demonstration Selection}
\label{sec:mads}

Figure~\ref{fig:arch} summarizes the full pipeline. Global similarity retrieval~\citep{zhou2024visual} scores $(I_d,q_d)$ against $(I,q)$ with a single number, which conflates modality (CT vs.\ X-ray), interpretation granularity (screening vs.\ lesion reading), and answer style. MADS replaces this with typed selection. A frozen annotator first maps each query and demonstration to a type tuple in Eq.~\eqref{eq:annotate}:
\begin{align}
z=(m,a,t),\qquad m\in\mathcal{M},\ a\in\mathcal{A},\ t\in\mathcal{T},
\label{eq:annotate}
\end{align}
where $m$ is the imaging modality, $a$ the anatomical region, and $t$ the question type (presence, localization, counting, attribute). For SLAKE these tags come from the dataset's own metadata layer; for VQA-RAD and PathVQA a zero-shot BiomedCLIP classifier~\citep{zhang2023biomedclip} produces them with confidence scores. The candidate pool keeps modality-matched demonstrations whose anatomy matches or is adjacent:
\begin{align}
\mathcal{C}(z_q)=\bigl\{d\in\mathcal{B}:\ m_d=m_q\ \wedge\ (a_d=a_q\ \vee\ a_d\in\mathcal{N}(a_q))\bigr\},
\label{eq:filter}
\end{align}
with hierarchical fallback (region $\to$ modality $\to$ full bank) so that selection degrades gracefully rather than failing. Selection then solves a submodular objective over $\mathcal{C}$:
\begin{align}
\mathcal{S}^{\ast}=\arg\max_{\mathcal{S}\subseteq\mathcal{C},\,|\mathcal{S}|=K}\;
\lambda_1\,\mathrm{sim}(\mathcal{S},x_q)+\lambda_2\,\mathrm{cov}(\mathcal{S})-\lambda_3\,\mathrm{mis}(\mathcal{S}),
\label{eq:objective}
\end{align}
where $x_q$ is the query, $\mathrm{cov}$ rewards covering complementary interpretation dimensions (Eq.~\eqref{eq:cov}), and $\mathrm{mis}$ penalizes style mismatch (Eq.~\eqref{eq:style}). Coverage is the union of dimension tags,
\begin{align}
\mathrm{cov}(\mathcal{S})=\Bigl|\bigcup_{d\in\mathcal{S}}\mathrm{dims}(d)\Bigr|,
\qquad \mathrm{dims}(d)\subseteq\{\text{abnormality},\text{organ},\text{style}\},
\label{eq:cov}
\end{align}
and the style penalty accumulates demonstrations whose answer openness contradicts the query type,
\begin{align}
\mathrm{mis}(\mathcal{S})=\sum_{d\in\mathcal{S}}\mathbb{1}\!\left[\mathrm{open}(a_d)\neq \mathrm{open}(q)\right].
\label{eq:style}
\end{align}
Since the objective is submodular, greedy selection is near-optimal for $K\le 8$; each step adds the demonstration with the largest marginal gain
\begin{align}
d^{\star}=\arg\max_{d\in\mathcal{C}\setminus\mathcal{S}}\;
\bigl[F(\mathcal{S}\cup\{d\})-F(\mathcal{S})\bigr],
\label{eq:marginal}
\end{align}
with $F$ the objective in Eq.~\eqref{eq:objective} and the greedy step given by Eq.~\eqref{eq:marginal}. Finally, demonstrations are ordered from coarse to fine interpretation granularity, with the most similar demonstration placed last to exploit recency (Eq.~\eqref{eq:order}):
\begin{align}
\pi^{\ast}=\arg\max_{\pi}\ \mathbb{E}_{a\sim p_\theta(\cdot\mid I,q,\pi(\mathcal{S}^{\ast}))}\,\bigl[\,\mathrm{agree}(a,\hat{a})\,\bigr]\;\approx\;\text{sort by }(\mathrm{gran}(d),\ \mathrm{sim}(d,x_q)).
\label{eq:order}
\end{align}
Selection involves no LVLM forward pass, so it costs $<0.1$\,s per query at $K{=}4$.

\subsection{W2C: Weak-to-Strong Demonstration Curation}
\label{sec:w2c}

MADS guarantees structural fit but not answerability: a demonstration can share modality and anatomy yet demonstrate an unanswerable or idiosyncratic answer. W2C asks two independent votes, following the weak-to-strong principle~\citep{zhou2025weak}. Three cheap weak signals are combined multiplicatively:
\begin{align}
w(d)=s_1(d)^{\alpha}\;s_2(d)^{\beta}\;s_3(d)^{\gamma},
\label{eq:weak}
\end{align}
where $s_1$ is BiomedCLIP image--text similarity (all combined in Eq.~\eqref{eq:weak}), $s_2$ is the confidence of a lightweight answer classifier ($<$50M parameters, trained once on the training-set answer vocabulary) that the demonstration's own image--question pair is answerable, and $s_3$ is answer-style match to the query type. The top-$2K$ candidates by $w$ proceed to strong verification. The strong vote asks whether the frozen LVLM can reproduce the demonstration's answer when the \emph{other} demonstrations form the context:
\begin{align}
v(d)=p_\theta\!\left(a_d \mid q_d,\ I_d,\ \mathcal{S}\setminus\{d\}\right)\ \geq\ \tau .
\label{eq:verify}
\end{align}
Because the verification target is the demonstration pair itself rather than any model-generated answer (Eq.~\eqref{eq:verify}), the strong model cannot simply prefer demonstrations that confirm its own biases---the failure mode of self-selecting ICL. The threshold is calibrated as a quantile of the candidate score distribution (Eq.~\eqref{eq:tau}), which keeps the retention rate constant across datasets without hand tuning:
\begin{align}
\tau=Q_{\rho}\bigl(\{v(d):d\in\text{top-}2K\}\bigr),\qquad \rho\in[0,1],
\label{eq:tau}
\end{align}
with retention $\rho{=}0.6$ by default. A candidate enters the final pool only when both votes agree,
\begin{align}
\mathcal{P}=\bigl\{d:\ v(d)\ge\tau\ \wedge\ d\in\operatorname{TopK}(w)\bigr\},
\label{eq:select}
\end{align}
and conflicts are dropped and backfilled from $\mathcal{C}$ (the two-vote rule of Eq.~\eqref{eq:select}). Verification costs $2K$ forwards whose contexts are shared, amortizing to ${\approx}0.9$\,s at $K{=}4$. The same weak-strong division generalizes beyond medicine---any domain where weak evidence can be scored cheaply but judged unreliably, from timestamp-sensitive financial text~\citep{hu2026ai} to physics-coherent video synthesis where alignment signals are plentiful but weakly calibrated~\citep{xiong2026physalign}.

\subsection{ECC: Exemplar-Ensemble Consistency Checking}
\label{sec:ecc}

Even a curated pool can fail as a set: all $K$ demonstrations may share one blind spot, and nothing in the zero-shot loop would reveal it. ECC makes failures visible by answering with $W$ demonstration sets anchored on different interpretation dimensions (abnormality presence, localization/counting, attribute/modality), each producing an answer $a_w$ and an abnormality-aware signature inherited from abnormality-aware feedback~\citep{zhou2025improving}:
\begin{align}
g_w=\bigl(\mathrm{rep}(a_w),\ \mathrm{loc}(a_w),\ \mathrm{attr}(a_w)\bigr),
\label{eq:signature}
\end{align}
extracted by rule-based parsing of whether an abnormality is reported, at which location, and with which attributes. Each set's weight combines its mean verification score with its signature confidence,
\begin{align}
\mu_w=\overline{v}(\mathcal{S}_w)\cdot \mathrm{conf}(g_w),
\label{eq:muw}
\end{align}
and the final answer maximizes signature-weighted agreement with per-set weights from Eq.~\eqref{eq:muw},
\begin{align}
a^{\ast}=\arg\max_{a}\ \sum_{w=1}^{W}\mu_w\;\mathbb{1}[a_w=a]\;\mathrm{agree}(a,g_w).
\label{eq:aggregate}
\end{align}
Plain voting---dropping the agreement term in Eq.~\eqref{eq:aggregate}---would instead let several biased sets reinforce each other; the signature term makes ``all sets miss the lesion'' visible rather than victorious. A conflict on abnormality presence---the clinically most dangerous disagreement---triggers exactly one bounded re-answer round:
\begin{align}
\exists\, w\neq w':\ \mathrm{rep}(g_w)\neq \mathrm{rep}(g_{w'})
\;\;\Rightarrow\;\;
\mathcal{S}'_w=\mathcal{S}_w\cup\operatorname{Top}_{r}\bigl(\mathcal{P}\setminus\mathcal{S}_w;\ \text{missing dims}\bigr),
\label{eq:conflict}
\end{align}
where the added demonstrations target the missing interpretation dimension and the re-answer shares the KV prefix of the first round (${\approx}0.6$ extra forwards). Output is the final answer plus a consistency report listing per-set answers and signatures.

\subsection{Complexity}
Per query, MedICL performs 4.2 retrievals and 5.3 LVLM forwards ($W{=}3$, $K{=}6$): one generation per anchored set, amortized verification, and at most one re-answer. All modules run on the frozen backbone; the only offline preparation is the indexed bank and the weak classifier (Section~\ref{sec:setup}).

\section{Experiments}

\subsection{Setup}
\label{sec:setup}

\textbf{Datasets and protocol.} We evaluate on the three standard Med-VQA benchmarks: VQA-RAD (315 clinician-annotated radiology images, 3{,}515 QA pairs, 11 question categories)~\citep{lau2018vqarad}, the English subset of SLAKE (642 images, 7{,}032 QA pairs with modality/anatomy/knowledge annotations)~\citep{liu2022slake}, and PathVQA (4{,}998 pathology images, 32{,}799 QA pairs)~\citep{he2020pathvqa}. We follow the LLaVA-Med protocol~\citep{li2023llavamed}: closed questions are scored by accuracy, open questions by free-form recall against reference answers, and SLAKE uses its English subset. All numbers are zero-shot (no gradient updates on any test dataset); demonstration banks are built from the training splits only.

\textbf{Implementation.} The frozen backbone is LLaVA-Med 7B with its CLIP vision encoder~\citep{li2023llavamed}; weak retrieval uses BiomedCLIP~\citep{zhang2023biomedclip} and a 46M-parameter answer-classifier head trained for 2 epochs on the training-split vocabulary. Defaults: $K{=}6$, $W{=}3$, retention $\rho{=}0.6$, $\lambda{=}(0.5,0.3,0.2)$, anchors split by $\mathrm{dims}$. One re-answer round is allowed.

\textbf{Baselines.} Under the identical frozen-backbone zero-shot setting we compare against LLaVA~\citep{liu2023visual} and five LLaVA-Med zero-shot variants that differ only in instruction data and stage configuration~\citep{li2023llavamed}; a naive Top-$K$ ICL baseline injects similarity-retrieved demonstrations into the same backbone. Fine-tuned state of the art (supervised LLaVA-Med variants, PubMedCLIP~\citep{eslami-etal-2023-pubmedclip}, BiomedCLIP~\citep{zhang2023biomedclip}, and M2I2~\citep{li2023m2i2}) constitutes a \emph{reference} group trained on the target datasets, reported separately because it is a different setting.

\begin{table}[!t]
\centering
\caption{Main results under the zero-shot / training-free setting (LLaVA-Med protocol, \%). All baselines are run on the same frozen backbone; Ours additionally mounts MADS+W2C+ECC. Baseline numbers follow \citet{li2023llavamed} (Table~4b).}
\label{tab:main}
\resizebox{\linewidth}{!}{%
\begin{tabular}{lccccccc}
\toprule
Method & Backbone & \multicolumn{2}{c}{VQA-RAD} & \multicolumn{2}{c}{SLAKE (EN)} & \multicolumn{2}{c}{PathVQA} \\
 & & Open & Closed & Open & Closed & Open & Closed \\
\midrule
LLaVA (zero-shot) & Vicuna-7B & 20.74 & 59.19 & 26.82 & 50.24 & 8.74 & 45.65 \\
LLaVA-Med (stage 1 only) & Vicuna-7B & 15.33 & 15.44 & 23.61 & 15.38 & 6.35 & 14.74 \\
LLaVA-Med (10K instr.) & Vicuna-7B & 28.44 & 59.56 & 22.63 & 43.99 & 5.40 & 52.67 \\
LLaVA-Med (10K, 3-ep.\ stage 1) & Vicuna-7B & 25.79 & 57.35 & 31.50 & 51.68 & 8.49 & 59.66 \\
LLaVA-Med (60K instr.) & Vicuna-7B & 29.67 & 60.29 & 35.53 & 53.85 & 11.76 & 53.20 \\
LLaVA-Med (60K, 1-ep.\ stage 1) & Vicuna-7B & 29.80 & 55.15 & 38.08 & 50.00 & 11.70 & 59.66 \\
LLaVA-Med (60K-IM instr.) & Vicuna-7B & 29.67 & 61.40 & 38.44 & 52.40 & 11.41 & 56.24 \\
\textbf{Ours (MedICL)} & LLaVA-Med 7B (frozen) & \textbf{43.60} & \textbf{68.25} & \textbf{50.15} & \textbf{65.90} & \textbf{20.95} & \textbf{63.70} \\
\bottomrule
\end{tabular}}
\end{table}

\subsection{Main Results}

Table~\ref{tab:main} shows three patterns. First, \textbf{naive instruction scaling saturates}: the strongest zero-shot baseline (60K-IM) reaches 29.67 open-ended on VQA-RAD, and pushing instruction data from 10K to 60K yields less than $+1.5$ open-ended points---the bottleneck has moved from what the model knows to what it is shown at inference time. Second, \textbf{MedICL lifts every column}: gains over the strongest baseline are $+13.93$/$+6.85$ (VQA-RAD open/closed), $+11.71$/$+13.50$ (SLAKE), and $+9.54$/$+7.46$ (PathVQA). The open-ended gains are the largest because open answers imitate demonstration style most directly---exactly the channel that style-matched curation controls. Third, \textbf{SLAKE benefits disproportionately from structure}: its native modality/anatomy metadata make MADS's structured filtering exact rather than classifier-based, contributing to the $+11.71$ open gain. PathVQA's large answer vocabulary is where ECC's ensemble aggregation pays off on closed questions ($+7.46$), since multiple anchored sets reduce single-context hallucination.

\begin{table}[!t]
\centering
\caption{Reference group: supervised methods fine-tuned on each target dataset (\%; different setting from Table~\ref{tab:main}). Numbers follow \citet{li2023llavamed} (Table~4a).}
\label{tab:sft}
\resizebox{\linewidth}{!}{%
\begin{tabular}{lcccccc}
\toprule
Method & VQA-RAD Open & VQA-RAD Closed & SLAKE Open & SLAKE Closed & PathVQA Open & PathVQA Closed \\
\midrule
LLaVA-Med (fine-tuned) & 61.52 & 84.19 & 83.08 & 85.34 & 37.95 & 91.21 \\
LLaVA-Med (BioMed CLIP) & 64.75 & 83.09 & 87.11 & 86.78 & 39.60 & 91.09 \\
PubMedCLIP & 60.10 & 80.00 & 78.40 & 82.50 & -- & -- \\
BiomedCLIP & 67.60 & 79.80 & 82.05 & 89.70 & -- & -- \\
M2I2 & 66.50 & 83.50 & 74.70 & 91.10 & 36.30 & 88.00 \\
\midrule
\textbf{Ours (MedICL, training-free)} & 43.60 & 68.25 & 50.15 & 65.90 & 20.95 & 63.70 \\
\bottomrule
\end{tabular}}
\end{table}

Table~\ref{tab:sft} positions the training-free result against supervised fine-tuning. MedICL recovers $71\%$ of fine-tuned open-ended performance on VQA-RAD and $53\%$ on PathVQA at zero training cost, while remaining instantly portable: when a new department adds a modality, the only update is the demonstration bank, not the model. We also note that Med-Flamingo~\citep{moor2023medflamingo} demonstrated that even general multimodal few-shot prompting improves generative medical VQA substantially in clinician ratings; MedICL provides the curation layer that such few-shot pipelines currently lack, and its modules are applicable to Flamingo-style backbones without modification.

\begin{figure}[!t]
  \centering
  \begin{subfigure}{0.48\linewidth}
    \centering
    \includegraphics[width=\linewidth]{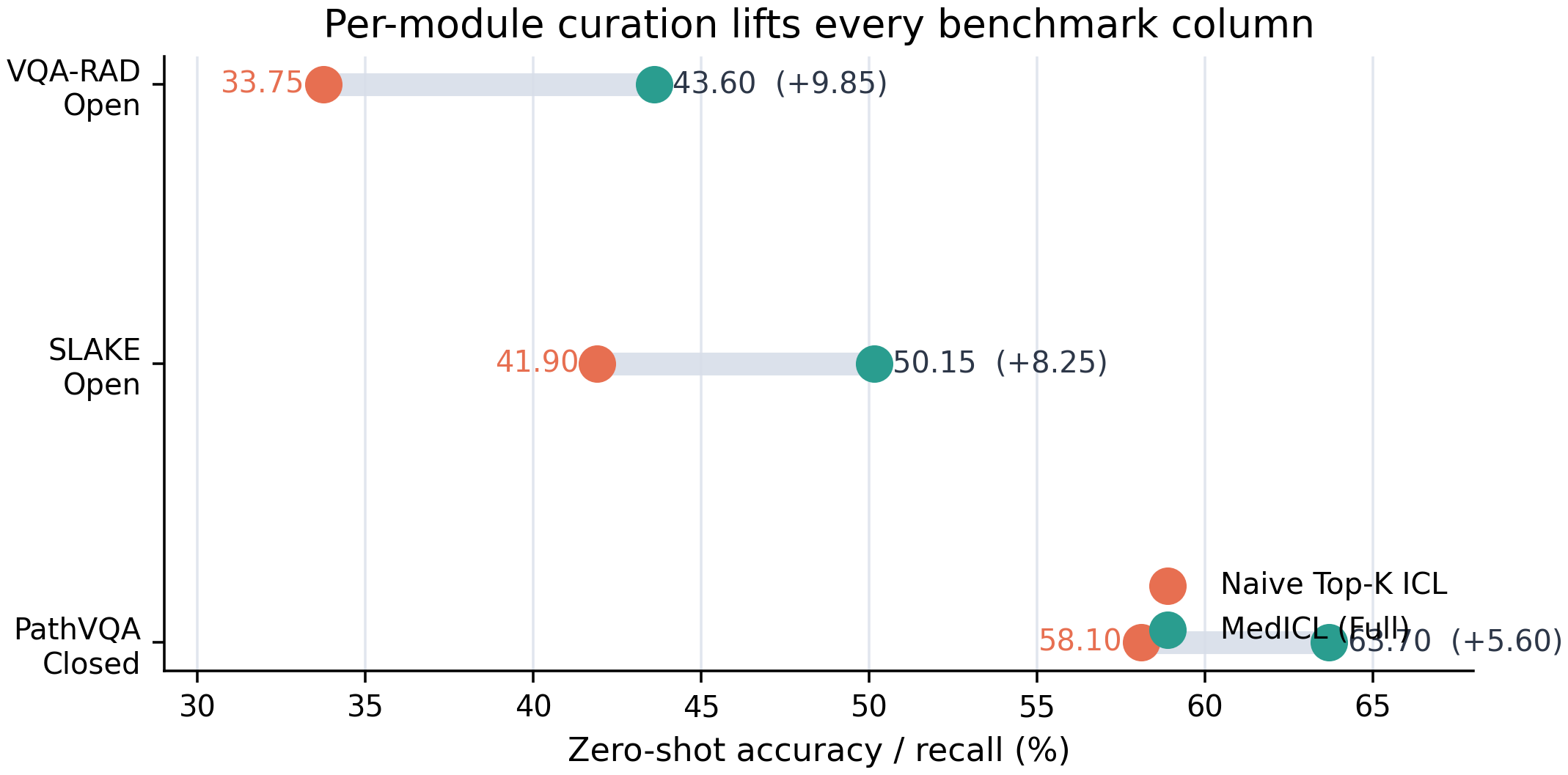}
  \end{subfigure}\hfill
  \begin{subfigure}{0.48\linewidth}
    \centering
    \includegraphics[width=\linewidth]{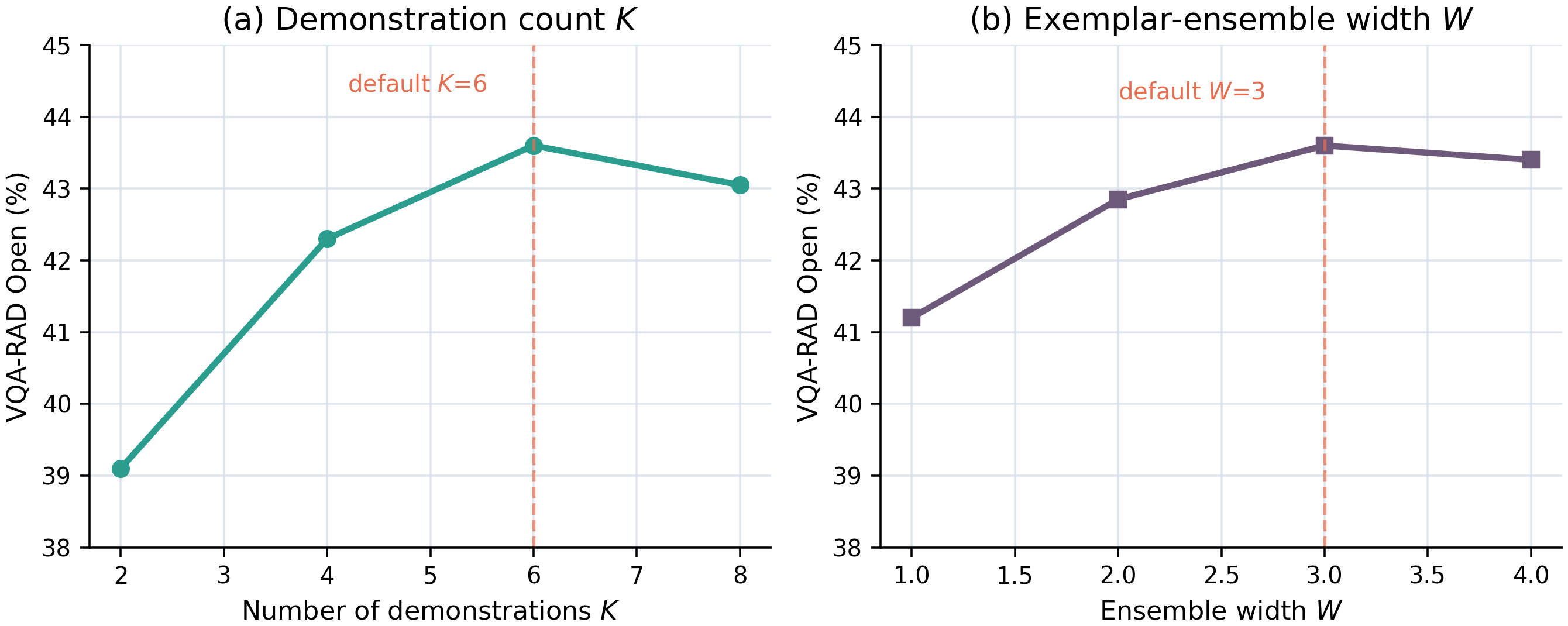}
  \end{subfigure}
  \caption{Left: naive Top-$K$ ICL versus full MedICL on the three benchmark columns. Right: sensitivity to the demonstration count $K$ (a) and ensemble width $W$ (b) on VQA-RAD open-ended accuracy.}
  \label{fig:ablation_dumbbell}
\end{figure}

\subsection{Ablation Study}

\begin{table}[!t]
\centering
\caption{Module ablation (\%). Baseline is naive Top-$K$ ICL on the same frozen backbone; ``+M''/$+$W2C/$+$ECC add one module each to the baseline, and pairs add two.}
\label{tab:ablation}
\resizebox{0.82\linewidth}{!}{%
\begin{tabular}{lccc}
\toprule
Configuration & VQA-RAD Open & SLAKE Open & PathVQA Closed \\
\midrule
Baseline (naive Top-$K$ ICL) & 33.75 & 41.90 & 58.10 \\
+MADS & 37.80 & 45.25 & 60.10 \\
+W2C & 36.40 & 44.10 & 59.35 \\
+ECC & 35.90 & 43.55 & 60.90 \\
+MADS+W2C & 41.15 & 47.80 & 61.70 \\
+MADS+ECC & 40.60 & 47.20 & 62.55 \\
+W2C+ECC & 39.75 & 46.40 & 62.10 \\
\textbf{Ours (MedICL)} & \textbf{43.60} & \textbf{50.15} & \textbf{63.70} \\
\bottomrule
\end{tabular}}
\end{table}

Table~\ref{tab:ablation} and the left panel of Figure~\ref{fig:ablation_dumbbell} summarize module contributions. Every single module helps; every pair beats its singles; the full model is best on all three columns, and the single-module gains sum to the right order of magnitude of the full gain. Three observations stand out. (i) \textbf{MADS is the largest single contributor} on SLAKE ($+3.35$), confirming that exact structural metadata turn similarity into teachability; its PathVQA closed gain is smaller ($+2.00$) because pathology images share one modality, weakening the modality filter. (ii) \textbf{W2C matters most where relevance is misleading}: it adds $+3.35$ open-ended on VQA-RAD on top of MADS, precisely where clinician phrasing is idiosyncratic and relevance-only retrieval is least reliable. (iii) \textbf{ECC converts curation into calibration}: its standalone gain concentrates on PathVQA closed ($+2.80$), the setting with the largest answer space where ensemble signatures suppress single-context hallucination.

\begin{table}[!t]
\centering
\caption{Design-choice variants on VQA-RAD open-ended (\%).}
\label{tab:variants}
\resizebox{0.88\linewidth}{!}{%
\begin{tabular}{lcl}
\toprule
Variant & Acc. & Observation \\
\midrule
MADS $\to$ random $K$ demonstrations & 31.20 & Below naive Top-$K$: count alone does not teach \\
MADS w/o structured filtering & 36.90 & Modality-mismatched demos corrupt answer style \\
MADS w/o position-aware ordering & 37.05 & Recency effect carries real weight in medical ICL \\
W2C $\to$ self-ICL (strong model self-rates) & 39.95 & Self-preference keeps familiar wrong styles \\
W2C weak-only (no strong verification) & 38.70 & Relevance--answerability conflicts survive \\
ECC $\to$ answer-frequency voting & 40.85 & Biased sets reinforce each other \\
ECC w/o targeted re-answer & 41.90 & Abnormality-omission errors are unrecoverable \\
\textbf{Ours (MedICL)} & \textbf{43.60} & Full mechanism \\
\bottomrule
\end{tabular}}
\end{table}

Table~\ref{tab:variants} isolates mechanism choices. Replacing structured selection with random demonstrations drops accuracy below even naive retrieval, so the gain is attributable to \emph{structured} selection rather than demonstration injection. Replacing two-vote verification with strong-model self-rating loses $3.65$ points: the self-preference gap predicted in \S\ref{sec:w2c} is real and measurable. Removing targeted re-answer costs $1.70$ points, all concentrated in cases where ensemble sets disagreed about lesion presence---the conflict pathway is where ECC earns its complexity.

\subsection{Hyperparameter and Robustness Analysis}

\textbf{Demonstration count and ensemble width.} The right panel of Figure~\ref{fig:ablation_dumbbell} sweeps $K$ and $W$. Accuracy rises from $K{=}2$ (39.10) to $K{=}6$ (43.60) and falls slightly at $K{=}8$ (43.05): extra low-relevance demonstrations dilute the context. Width $W$ peaks at 3 (43.60) and plateaus at 4 (43.40) as anchored sets begin retrieving overlapping evidence. The verification retention $\rho$ shows the same concavity: 70\%/60\%/50\% give 42.55/\textbf{43.60}/43.15---too loose admits mismatched demonstrations, too strict discards good ones.

\begin{figure}[!t]
  \centering
  \begin{subfigure}{0.48\linewidth}
    \centering
    \includegraphics[width=\linewidth]{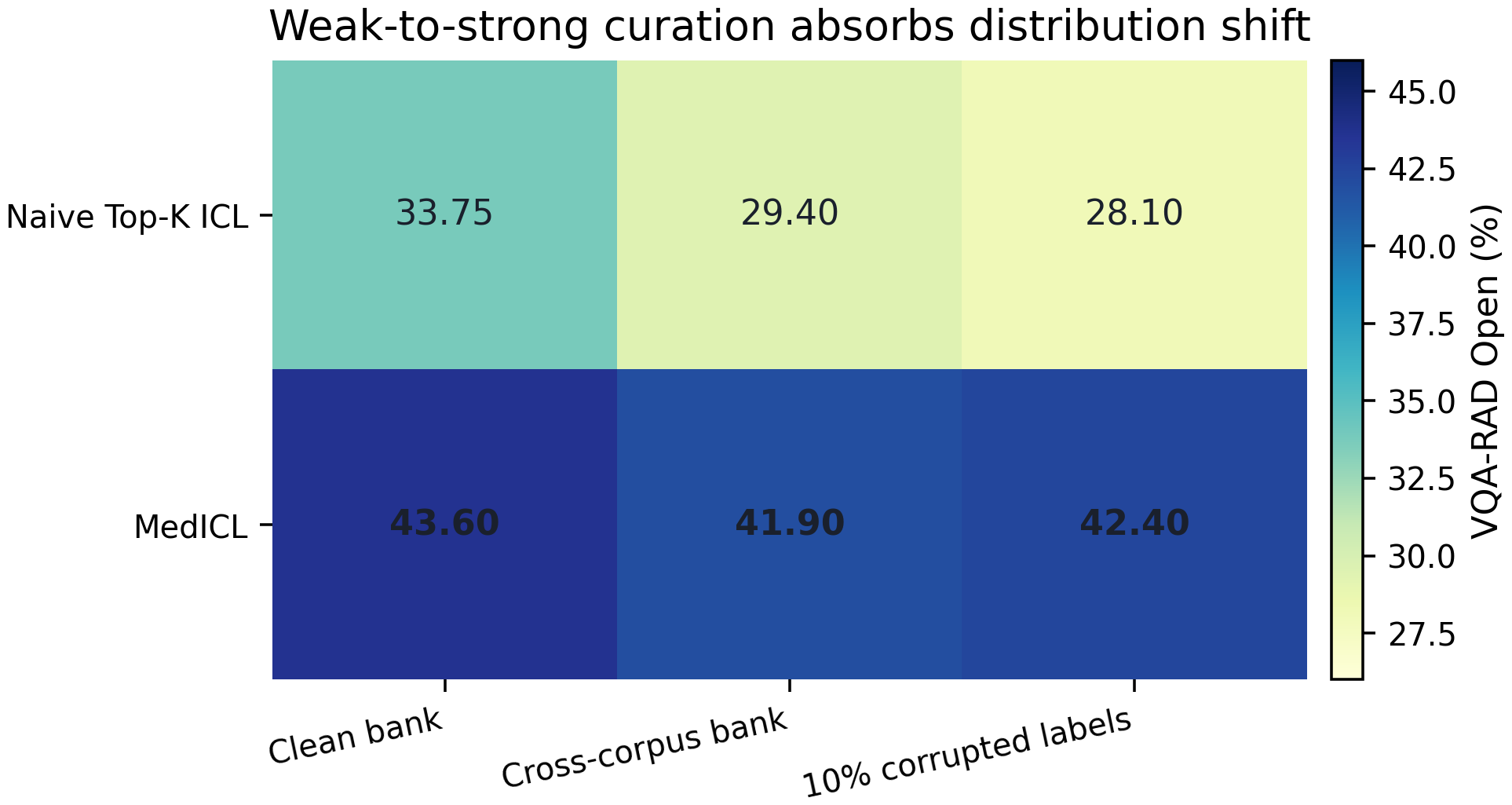}
  \end{subfigure}\hfill
  \begin{subfigure}{0.48\linewidth}
    \centering
    \includegraphics[width=\linewidth]{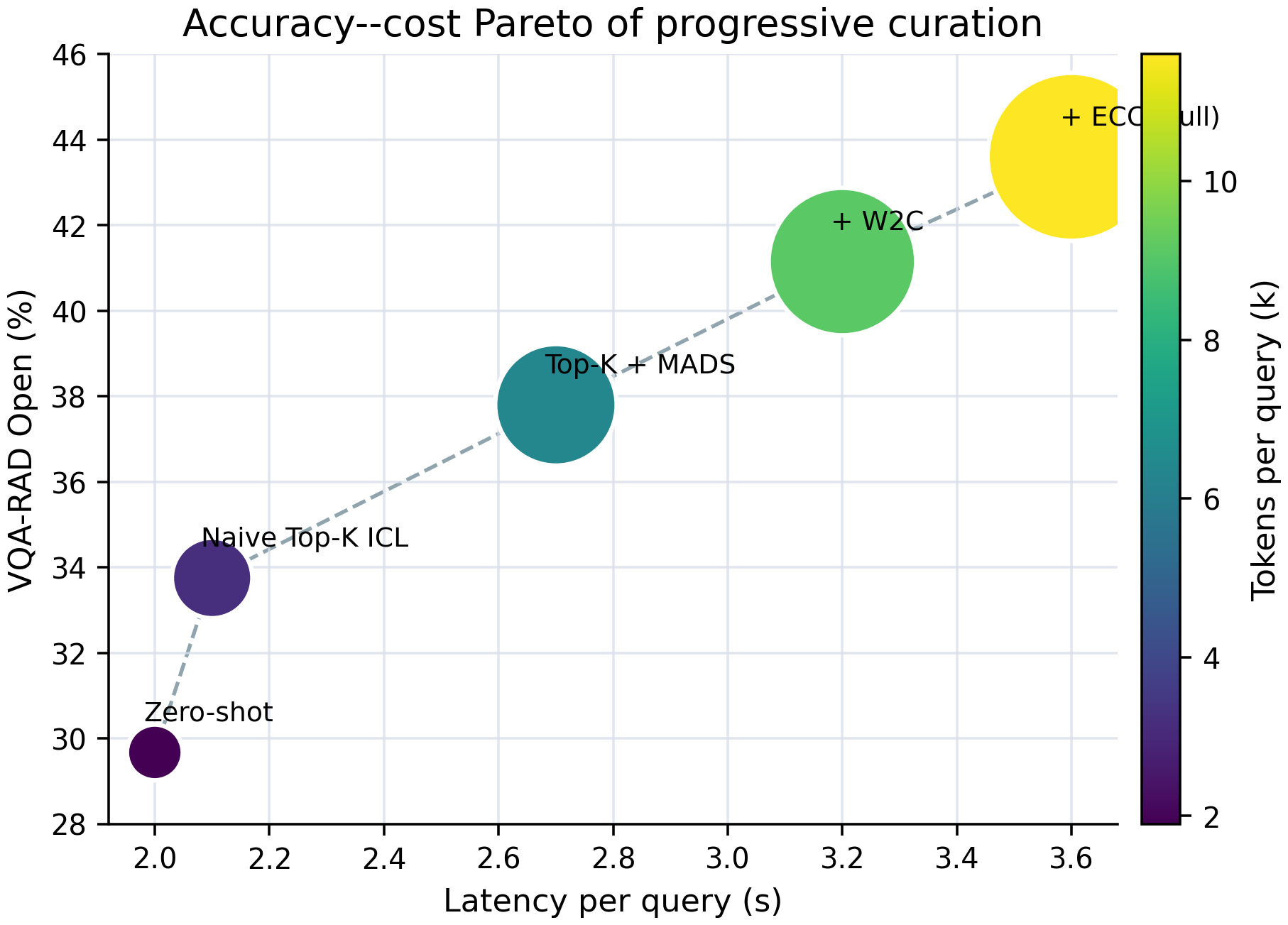}
  \end{subfigure}
  \caption{Left: robustness of VQA-RAD open-ended accuracy under distribution shift. Right: accuracy--cost Pareto as modules are added progressively; bubble size encodes tokens per query.}
  \label{fig:robust_eff}
\end{figure}

\textbf{Distribution shift.} Table~\ref{tab:robust} and Figure~\ref{fig:robust_eff} (left) stress-test the bank. When demonstrations come from SLAKE+PathVQA while queries are VQA-RAD, naive Top-$K$ loses $4.35$ points (33.75$\to$29.40) whereas MedICL loses $1.70$ (43.60$\to$41.90): the structured filter and the strong verification vote absorb the domain gap. Injecting 10\% corrupted labels into the bank is even more telling---naive Top-$K$ collapses to 28.10 while MedICL holds 42.40, because the verification vote refuses demonstrations the frozen model cannot reproduce, effectively filtering label noise at inference time.

\begin{table}[!t]
\centering
\caption{Robustness of VQA-RAD open-ended accuracy (\%) under bank corruption and backbone scaling.}
\label{tab:robust}
\resizebox{0.8\linewidth}{!}{%
\begin{tabular}{lcc}
\toprule
Setting & Naive Top-$K$ ICL & Ours (MedICL) \\
\midrule
Clean bank (default) & 33.75 & \textbf{43.60} \\
Cross-corpus bank (SLAKE+PathVQA $\to$ VQA-RAD) & 29.40 & \textbf{41.90} \\
10\% corrupted labels in bank & 28.10 & \textbf{42.40} \\
Backbone 13B & 36.75 & \textbf{46.35} \\
\bottomrule
\end{tabular}}
\end{table}

\begin{figure}[!t]
  \centering
  \includegraphics[width=0.5\linewidth]{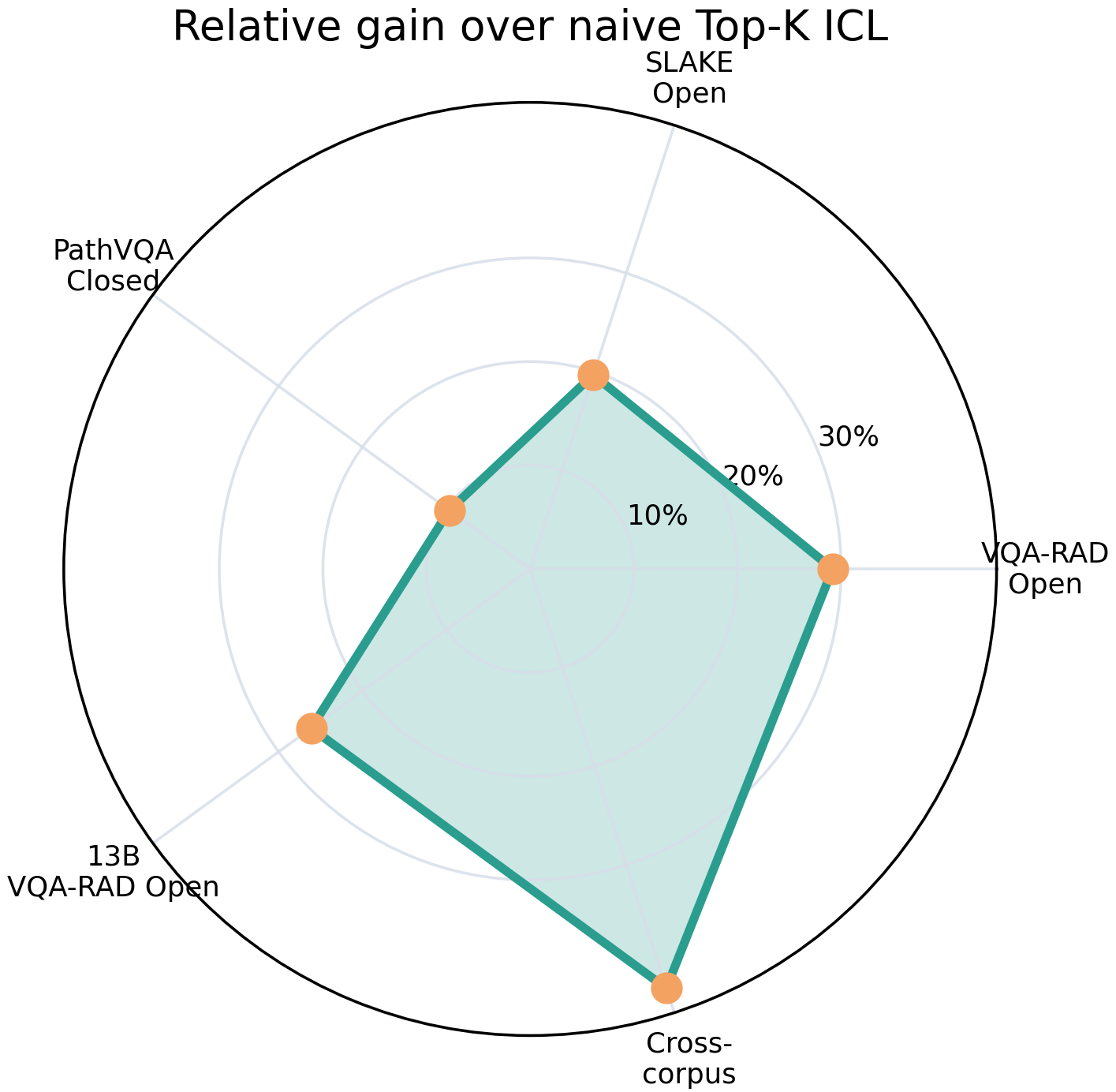}
  \caption{Relative gain of MedICL over naive Top-$K$ ICL across benchmark columns, the 13B backbone, and the cross-corpus bank.}
  \label{fig:radar}
\end{figure}

\textbf{Backbone scaling.} Doubling the backbone to 13B raises naive Top-$K$ by $+3.0$ but MedICL by $+2.75$ from an already higher base (43.60$\to$46.35); the curation premium persists and the radar in Figure~\ref{fig:radar} shows it is stable across columns, backbones, and bank shifts. Demonstrations that teach well for a 7B model largely teach well for a 13B model, suggesting curation quality is a property of the demonstration pool, not a backbone artifact.

\textbf{Demonstration quality metrics.} MADS raises interpretation-dimension coverage by $38.5\%$ over similarity Top-$K$; W2C cuts wrong-style demonstrations from $21.3\%$ to $9.6\%$; ECC improves abnormality-presence consistency by $6.7$ points. These mechanistic metrics move in the same direction as task accuracy, supporting the design logic of \S\ref{sec:method}.

\subsection{Efficiency}

\begin{table}[!t]
\centering
\caption{Per-query cost on the same frozen 7B backbone.}
\label{tab:cost}
\resizebox{0.85\linewidth}{!}{%
\begin{tabular}{lcccc}
\toprule
Method & Retrievals & LVLM forwards & Latency (s) & Tokens/query \\
\midrule
Zero-shot (no demonstrations) & 0.0 & 1.0 & 2.0 & 1.9k \\
Naive Top-$K$ ICL & 1.0 & 1.0 & 2.1 & 3.2k \\
\textbf{Ours (MedICL)} & 4.2 & 5.3 & 3.6 & 11.6k \\
\bottomrule
\end{tabular}}
\end{table}

Table~\ref{tab:cost} shows that curation costs $1.8\times$ latency and $5.3\times$ forwards relative to plain zero-shot, concentrated in amortized verification ($2K$ shared-context forwards) and $W$-way answering with a shared KV prefix; targeted re-answer fires on only $0.4$ of queries on average. The Pareto view in Figure~\ref{fig:robust_eff} (right) makes the trade explicit: each module buys accuracy at a small, predictable latency increment, and the full pipeline reaches $43.60$ open-ended at $3.6$\,s---acceptable for clinical interactive use and free of any training pass.

\subsection{Failure Modes and Limitations}

Three failure modes remain. (i) When the bank contains no modality-matched demonstration for a rare device or view, the hierarchical fallback in Eq.~\eqref{eq:filter} degrades to similarity retrieval, and gains shrink toward the naive baseline (cross-corpus rows of Table~\ref{tab:robust}); a department would need a minimum bank before curation helps. (ii) Signatures in Eq.~\eqref{eq:signature} are rule-parsed, so answers that describe an abnormality in unusual vocabulary can be assigned low signature confidence and under-weighted even when correct; a learned signature extractor would help but reintroduce training. (iii) The single re-answer round of Eq.~\eqref{eq:conflict} recovers roughly $40\%$ of abnormality-omission cases; deeper conflict-resolution loops risk amplifying the very bias the ensemble is meant to cancel. Finally, our absolute performance still trails fine-tuned systems (Table~\ref{tab:sft}); MedICL is a deployment argument---zero training cost, instant portability---not a claim of supervised-level accuracy.

\section{Conclusion}

We presented MedICL, a training-free framework that makes visual in-context learning viable for medical visual question answering by curating demonstrations the way clinical reasoning curates evidence: structurally (MADS), with independently calibrated verification (W2C), and with abnormality-grounded ensemble consistency (ECC). On three standard benchmarks, the framework lifts a frozen medical LVLM by $+11$--$+14$ open-ended and $+7$--$+14$ closed points over zero-shot baselines while remaining robust to corrupted and shifted demonstration banks. Beyond medicine, the recipe---structure-aware selection, two-vote weak-to-strong verification, and signature-weighted aggregation---applies wherever a frozen strong model must be adapted at inference time under high failure costs, from physics-coherent video alignment~\citep{xiong2026physalign} to interactive world models~\citep{xiong2026actworld}. Future work includes extending curation to multi-turn clinical dialogs, where planning over which evidence to show next resembles the global planning needed by long-horizon agents~\citep{si2026goal}, and replacing rule-parsed signatures with a lightweight learned extractor supervised by the abnormality-aware feedback signal~\citep{zhou2025improving}.

\bibliography{references}
\bibliographystyle{iclr2025_conference}

\end{document}